\documentclass[letterpaper,10pt,conference]{ieeeconf}

\IEEEoverridecommandlockouts
\usepackage{graphicx}
\usepackage{amsmath}
\usepackage{amssymb}
\usepackage{bm}
\usepackage{booktabs}
\usepackage{cite}
\usepackage{url}
\usepackage{capt-of}

\usepackage{xcolor}
\usepackage{pifont}

\definecolor{tickgreen}{RGB}{0,145,85}
\definecolor{crossred}{RGB}{205,45,45}

\newcommand{\cmark}{\textcolor{tickgreen}{\ding{51}}}
\newcommand{\xmark}{\textcolor{crossred}{\ding{55}}}

\providecommand{\Log}{\mathrm{Log}}
\providecommand{\clip}{\mathrm{clip}}

\providecommand{\argmin}{\operatorname*{arg\,min}}
\newcommand{\meanstd}[2]{#1\,{\tiny$\pm$}\,#2}
\newcommand{\bestmeanstd}[2]{\textbf{#1}\,{\tiny$\boldsymbol{\pm}$}\,\textbf{#2}}

\title{\LARGE \bf
TERRA: Terrain-Aware Reconstruction, Retargeting and Control for Musculoskeletal Locomotion
}

\author{Merkourios Simos$^{1}$, Chengkun Li$^{1}$, Bianca Ziliotto$^{1}$, Alexander Mathis$^{1}$
\thanks{$^{1}$EPFL}%
\thanks{Contact: {\tt\small alexander.mathis@epfl.ch}}%
\thanks{This work was supported by Swiss National Science Foundation (SNSF) (310030\_212516), the Simons foundation (SFI-AN-NC-SCN-00007276-14), and a Boehringer Ingelheim Fonds PhD stipend (B.Z.). M.S. acknowledges the support of the Onassis Foundation Scholarships' Program.}}

\IEEEaftertitletext{%
  \begin{minipage}{\textwidth}
    \centering
    \includegraphics[width=\linewidth]{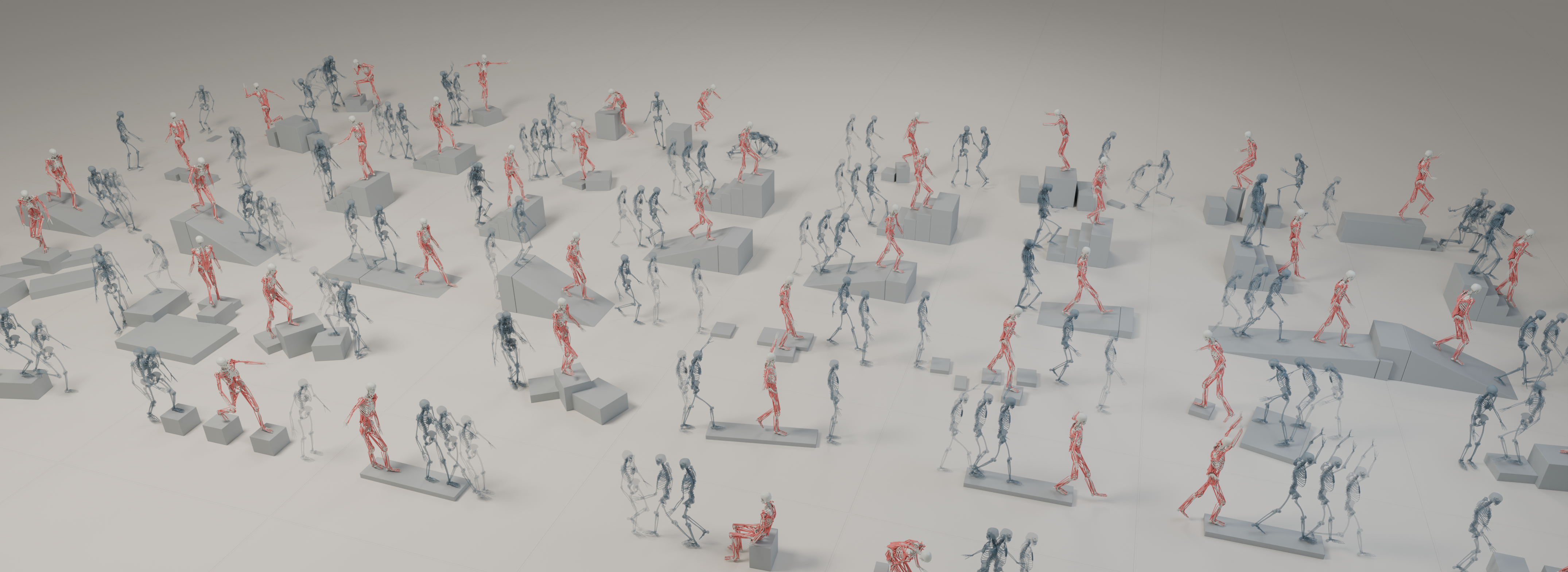}%
    \captionof{figure}{Representative outputs from TERRA. Each tile shows a
    motion-derived static collision terrain and the corresponding MyoFullBody
    reference used for muscle-driven tracking. The supported terrain families (affordances) are
    ramps, staircases, independent horizontal supports, and seats.}
    \label{fig:overview}
  \end{minipage}%
}

\begin{document}

\maketitle
\thispagestyle{empty}
\pagestyle{empty}

\begin{abstract}
Recent advances in musculoskeletal modeling and reinforcement learning have enabled muscle-actuated agents to reproduce increasingly complex human motions. Yet these capabilities remain largely confined to flat ground, in part because motion datasets rarely include aligned terrain geometry and because retargeting terrain interactions to complex musculoskeletal bodies is challenging.
We present TERRA, an end-to-end pipeline for terrain-aware retargeting and control of musculoskeletal locomotion. From kinematic trajectories alone, TERRA combines terrain priors, estimated contacts, and negative free-space evidence to recover task-relevant support geometry. TERRA further considers anatomical, tendon-continuity, and contact constraints during retargeting. Using the resulting motion-terrain pairs from five datasets, we successfully train a single muscle-actuated control policy on 9.4 hours of diverse locomotion. Across reconstruction, retargeting, and held-out tracking benchmarks, TERRA improves terrain accuracy, sharply reduces anatomical and interaction violations, and achieves the highest observed completion rate over supported terrain families. Overall, TERRA provides a practical route from scene-less motion data to muscle-actuated locomotion over diverse non-flat terrain. Project website: https://cnai.epfl.ch/terra/
\end{abstract}

\section{Introduction}
\label{sec:introduction}

Whether trail-running on Mont Blanc, climbing a flight of stairs, or simply sitting down in a chair, humans need to coordinate hundreds of muscles as the environment and the underlying terrain change.
Understanding how such control arises across diverse affordances~\cite{gibson1979ecological} is a fundamental problem in neuroscience~\cite{angelaki2026simons} and an increasingly practical one in robotics, as humanoids must navigate the same spaces designed for humans.

Physics-based musculoskeletal models combined with reinforcement learning (RL) provide a powerful framework for studying how complex movement can emerge from muscle-level control.
Early work used task-driven objectives to generate individual skills such as walking and running~\cite{wang2012optimizing,lee2014locomotion}.
More recent approaches leveraged large-scale motion-capture datasets and motion imitation to learn broad behavioral repertoires that can be reused for downstream tasks~\cite{feng2023musclevae,simos2025kinesis,li2026musclemimic,wang2026myochallenge,wei2026scaling}.
However, the behavioral repertoires remain almost entirely confined to flat ground.
Two challenges make non-flat locomotion particularly difficult.
First, existing motion-capture datasets rarely provide aligned terrain geometry.
Second, transferring human motion to a complex musculoskeletal body requires more than matching joint positions: the retargeted motion must preserve contacts, avoid penetration, foot skating, and floating, and remain compatible with anatomical joint and musculotendon limits.
These errors are amplified by non-flat terrain interaction and can render a reference motion infeasible (Table~\ref{tab:method-comparison}).

We present \textbf{TERRA}, an end-to-end framework for \textbf{terra}in-aware musculoskeletal retargeting and control.
From kinematics alone, TERRA reconstructs motion-relevant support geometry by combining terrain priors with estimated contact events, foot orientation, and free-space evidence from the moving body.
It then extends interaction-mesh retargeting with anatomical, contact, clearance, and collision constraints to produce valid references for a full-body, muscle-actuated model.
The resulting motion--terrain pairs form a 9.4-hour, five-source library for a
single reference-conditioned controller~\cite{mahmood2019amass,grimmer2023darmstadt,hori2026grip,vielemeyer2026ramp,boo2025gait120}.
Across ramps, stairs, platforms, and seats, TERRA reconstructs
motion-supported terrain, reduces reference violations, and enables one controller to
complete held-out motions from all four families.
Lastly, we qualitatively compare generated muscle activity with held-out EMG and vertical ground-reaction-forces (GRF) across flat ground, ramps, stairs, and sit/stand motions, characterizing similarities and differences in normalized waveform shape. Code and data will be made publicly available.

\section{Related Work}
\label{sec:related_work}

\textbf{Motion-conditioned terrain reconstruction.}
Several recent methods recover or synthesize surrounding scene structure based on motion trajectories, using learned contact and free-space cues or physics-based interaction constraints~\cite{yi2023mime,li2024physics}.
Complementary video-based pipelines jointly reconstruct human motion and scene geometry but rely on visual observations~\cite{allshire2025visual,wang2026contact,zhang2026meshmimic}.
Without visual scene observations, TIP jointly estimates inertial motion and a local terrain height field~\cite{jiang2022tip}.
Most closely related, SceneBot constructs contact-rich environments from retargeted motion by placing and merging constant-height terrain patches~\cite{chen2026scenebot}.
TERRA instead performs structured inference over explicit terrain priors, including platforms, stair flights, inclined ramps, and seated supports.
Whereas SceneBot evaluates its reconstructed scenes primarily through downstream tracking success, TERRA additionally measures geometry against paired ground-truth terrain. Because SceneBot's code is not publicly available, we compare against an explicit constant-height patch baseline adapted from its method and TIP.

\textbf{Motion retargeting.}
Large-scale human-motion repositories have made motion imitation a viable and scalable strategy for training general-purpose humanoid controllers~\cite{mahmood2019amass,bones_seed,lee2025phuma}.
Recorded motions must first be mapped into a robot's morphology. Existing methods include keypoint-based optimization supporting diverse embodiments~\cite{araujo2026gmr}, and learned cross-morphology mappings~\cite{huang2026human2humanoid}.
Recent work has also incorporated physical feasibility~\cite{lee2025phuma,wang2026spark}. Lastly, OmniRetarget proposed a shared body-scene interaction mesh that preserves interactions with objects and terrain~\cite{yang2026omniretarget}.
While OmniRetarget assumes access to the terrain or object geometry used to construct that interaction mesh, TERRA addresses a complementary upstream problem: it recovers
task-relevant support geometry when a motion dataset provides body kinematics without a scene model. TERRA adapts OmniRetarget's interaction-mesh representation and adds
target-specific anatomical and terrain-interaction terms for a muscle-driven
embodiment.

\textbf{Musculoskeletal control.}
Advances in musculoskeletal modeling and simulation have made physiologically detailed, contact-rich bodies increasingly accessible to learning-based control~\cite{lee2019scalable,seth2011opensim,geijtenbeek2019scone,caggiano2022myosuite}.
Several works tackled the challenge of efficient exploration in high-dimensional muscle actuation space~\cite{chiappa2023latent,he2024dynsyn,wei2026scalable}, enabling multi-task control of increasing dexterity and even athleticism~\cite{schumacher2025natural,an2026arnold,wang2026myochallenge, wang2026learning}.
More recently, motion imitation has enabled reusable locomotor repertoires, empirical validation of simulated muscle activity, and universal policies controlling muscle-actuated bodies across hundreds of reference motions~\cite{feng2023musclevae,simos2025kinesis,li2026musclemimic,wei2026scaling}.
Nevertheless, musculoskeletal imitation remained largely confined to flat ground.
TERRA extends musculoskeletal imitation learning to three-dimensional locomotion by training a single muscle-actuated policy on motions paired with reconstructed terrain.

\begin{table}[t]
  \centering
  \footnotesize
  \setlength{\tabcolsep}{3pt}
  \renewcommand{\arraystretch}{1.12}
  \caption{Capabilities of representative motion-processing methods. MSK denotes
  support for musculoskeletal target models.}
  \vspace{-5pt}
  \label{tab:related_capabilities}
  \resizebox{\columnwidth}{!}{%
    \begin{tabular}{@{}lcccc@{}}
      \toprule
      Method
        & \shortstack{Kinematic\\constraints}
        & \shortstack{Terrain\\reconstruction}
        & \shortstack{Terrain\\retargeting}
        & \shortstack{MSK} \\
      \midrule
      GMR~\cite{araujo2026gmr}
        & Joint bounds
        & \xmark & \xmark & \xmark \\
      OmniRetarget~\cite{yang2026omniretarget}
        & Hard interaction
        & \xmark & \cmark & \xmark \\
      SceneBot~\cite{chen2026scenebot}
        & Joint bounds
        & \cmark & \cmark & \xmark \\
      MuscleMimic~\cite{li2026musclemimic}
        & Joint/equality
        & \xmark & \xmark & \cmark \\
      \textbf{TERRA}
        & \textbf{Hard + anatomy}
        & \cmark & \cmark & \cmark \\
      \bottomrule
    \end{tabular}%
  }
  \label{tab:method-comparison}
  \vspace{-16pt}
\end{table}

\begin{figure*}[t!]
  \centering
    \includegraphics[width=1\textwidth]{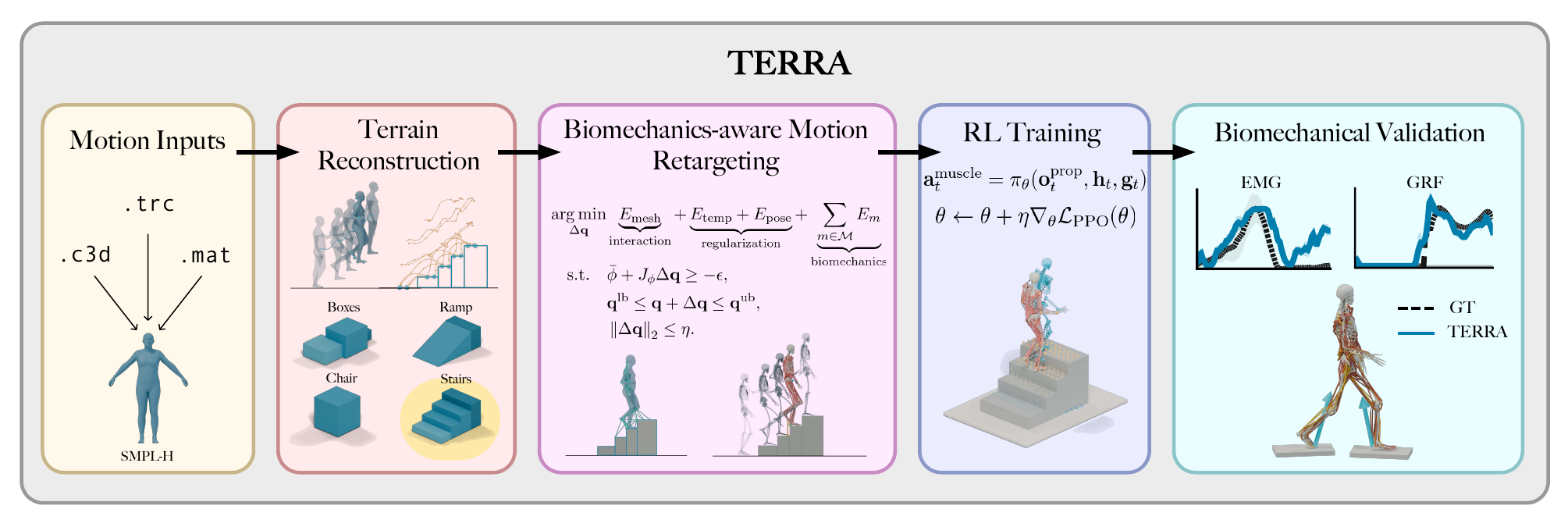}%
  \vspace{-13pt}
  \caption{\textbf{TERRA method overview.} Heterogeneous motion data are converted to
  SMPL-H, used to reconstruct compact collision terrain, and retargeted to the
  musculoskeletal model with terrain-aware constraints. The resulting paired
  trajectory--terrain scenes train a muscle-actuated tracking policy, whose
  biomechanical outputs are compared with subject-matched force-plate and EMG data
  on held-out motions.}
  \label{fig:method-overview}
  \vspace{-15pt}
\end{figure*}

\section{Method}
\label{sec:method}

We first introduce the musculoskeletal model, source motion, and terrain priors.
We then describe the terrain reconstruction method, the
retargeting stage, musculoskeletal controller training, and the  biomechanical comparison pipeline~(Fig.~\ref{fig:method-overview}).

\subsection{Problem Setup}
\label{sec:setup}

\textbf{Musculoskeletal model.}
We used the finger-disabled MyoFullBody configuration from
MuscleMimic~\cite{li2026musclemimic}, with 354 Hill-type musculotendon actuators and 83 MuJoCo joints: one floating-root joint and 82 scalar articulated joints. The
model also contains 49 active polynomial couplings for dependent lumbar,
scapulohumeral, and knee joints.

In order to align the model with reference motion data, we paired $K=17$ SMPL-H~\cite{romero2017embodied} landmarks with MyoFullBody body origins: pelvis, spine, and
head; bilateral hip, knee, ankle, and toe; and bilateral shoulder, elbow, and
wrist. We denote their target-model positions by $\bm p_i(\bm q)$. Alignment was performed by placing
MyoFullBody and SMPL-H in corresponding neutral poses, optimizing the SMPL-H
shape coefficients and global scale, and applying residual position offsets to source motions. Collision
geometries in the feet, thighs, shanks, and posterior pelvis were
modeled as capsule and ellipsoid primitives.

\textbf{Source motion.}
TERRA accepts motion inputs in AMASS-compatible SMPL-H format~\cite{mahmood2019amass}.
For marker-based datasets, TERRA additionally supports C3D, MAT, and TRC formats~(Fig.~\ref{fig:method-overview}). Each sequence is transformed into 52 world-space
joint positions and rotations using the SMPL-H shape fitted to MyoFullBody, then
translated vertically so that the lowest fitted toe lies at $z=0$.

\textbf{Terrain priors.}
TERRA defines four terrain priors: independent support boxes, inclined ramps,
staircases, and seat supports (Fig.~\ref{fig:method-overview}).
Each prior is implemented as an assembly of finite, static MuJoCo box collision
geometries. Independent supports and
seats use horizontal boxes, staircases use an ordered set of horizontal boxes, and ramps use a pitched box with a planar top face.
All terrain boxes use the same contact and friction
parameters as the floor.

\subsection{Terrain Reconstruction}
\label{sec:terrain}

TERRA constructs terrain directly from motion trajectories via terrain priors.
Intervals in which an ankle or toe landmark remains locally stationary and low provide
\emph{positive surface evidence}, while the remaining body trajectory provides
\emph{free-space evidence} that bounds the admissible geometry. TERRA distinguishes
discrete horizontal surfaces from continuous ramps using two additional motion
cues: the neutral-calibrated ankle--toe orientation during stance and the
swing-foot clearance profile between consecutive stance intervals of the same foot.

\textbf{Support extraction and calibration.}
For each left and right ankle and toe landmark $j$, we define candidate stationary
intervals $e$ as contiguous runs satisfying
$\lVert\dot{\bm x}_{j,t}\rVert_2<v_{\mathrm s}$ for all $t\in e$. A candidate is
retained when
\begin{equation}
 |e|\geq n_{\min},\qquad
 \bar z_e\leq\min_{\tau\in\mathcal W_e}z_{j,\tau}+\epsilon_{\mathrm{loc}},
 \label{eq:terrain_stance}
\end{equation}
where $\bar z_e$ is the median landmark height and $\mathcal W_e$ is a local temporal window.
Candidate ankle and toe intervals that overlap with a detected seated interval are excluded from surface-height estimation.

To account for the distance between the joint centers and the body's contact surface, we subtract a joint-specific offset
$o_j$ and treat $\hat h_e=\bar z_e-o_{j(e)}$ as surface height. When available, offsets come from
a subject-matched flat reference; otherwise, we estimate them from self-calibration for
datasets with a flat portion, or from the lowest motion-supported level.
Distinct surface-height levels are computed by sorting
$\hat h_e$ and starting a new level whenever adjacent values differ by more than
$4$\,cm. Any initial group spanning more than $8$\,cm is recursively divided at
its largest internal gap. The height of each level is computed by first averaging
$\hat h_e$ within each landmark (e.g. left toe) and then averaging across landmarks.

\textbf{Continuous versus discrete terrain.}
Let $u$ denote the projection of each stationary interval's median
horizontal location onto the first principal direction $\bm a$ of these
locations, oriented toward increasing height. We fit the flat--incline--flat profile
\begin{equation}
 h(u)=h_0+s\,[\clip(u,u_0,u_1)-u_0]
 \label{eq:ramp_profile}
\end{equation}
where $h_0$ is the lower landing height, $s$ is the incline grade, and $u_0$ and
$u_1$ are the coordinates at which the incline begins and ends. Let $\theta_t$
be the ankle--toe pitch, $\theta_\ell^0$ its neutral value for foot $\ell$, and
$\gamma_t$ the angle between the foot heading and ramp direction. Using medians
first within each stationary interval and then across intervals, we compute
\begin{equation}
 \begin{aligned}
 E_{\mathrm{flat}}&=\operatorname{med}|\theta_t-\theta_\ell^0|,\\
 E_{\mathrm{ramp}}&=\operatorname{med}
 |\theta_t-\theta_\ell^0-\arctan(s\cos\gamma_t)|.
 \end{aligned}
 \label{eq:foot_orientation}
\end{equation}
The smaller error determines whether the foot orientation better matches a flat
or inclined surface. When
$|E_{\mathrm{flat}}-E_{\mathrm{ramp}}|\leq2^\circ$, a normalized
swing-clearance score compares the foot trajectory with a linear height
transition and assigns near-linear trajectories to ramps and trajectories with
early ascent or delayed descent to discrete surfaces. The $2^\circ$ ambiguity
band corresponds to approximately 7-mm endpoint uncertainty over a 20-cm foot
chord; the swing cue must exceed 5\% of the observed support-level change.

\textbf{Discrete terrain geometry.}
For motions containing at least three raised surface-height levels, a staircase
candidate with a common horizontal axis and width with fitted
riser height is considered. An independent-box candidate is also constructed. The yaw
of each raised level is aligned with the principal axis of its horizontal contact
points, and spatially separated contacts on the same level are assigned to
different boxes. Each contact envelope is expanded by a foot-sized margin, then
trimmed where the resulting geometry would intersect free-space body
trajectories. A staircase prior is selected when
candidate treads are within $5$\,cm of every raised stationary foot and no other
body landmark penetrates it by more than $5$\,cm; otherwise, terrain is
constructed with independent boxes.

\textbf{Seat reconstruction.}
Seats are represented by boxes and detected when pelvis speed remains below $0.15$\,m/s for at
least $0.4$\,s and its median horizontal position lies at least $0.10$\,m outside
the convex hull of the feet and grounded limb landmarks. The seat top is the
median of the $10$th-percentile heights of the robot-fitted SMPL-H posterior
pelvis/hip surface at the first, middle, and last interval frames. Candidates outside the $0.04$--$0.60$\,m
height range are discarded.

\vspace{-3pt}
\subsection{Terrain-Aware Motion Retargeting}
\label{sec:retargeting}

Once a terrain is constructed, TERRA retargets the paired motion by adapting
OmniRetarget's interaction-mesh objective, adding target-specific anatomical and
terrain-interaction terms, and solving the resulting objective by sequential quadratic
programming.

\textbf{Interaction mesh.}
Following OmniRetarget~\cite{yang2026omniretarget}, let $\bm x_{i,t}$ be the
source landmark corresponding to the target-model position $\bm p_i(\bm q)$,
and let $S=[\bm s_1,\ldots,\bm s_{N_S}]^\top$ contain samples of the
reconstructed scene.
We sample each box top with spacing no greater than $0.20$\,m and add a coarse floor grid after removing points inside box footprints.
At each frame, the source and robot vertex matrices are
\vspace{-2pt}
\begin{equation}
\begin{aligned}
\widetilde V_t
 &= \bigl[\bm{x}_{1,t}^{\mathsf T}\ \cdots\
          \bm{x}_{K,t}^{\mathsf T}\ S^{\mathsf T}\bigr]^{\mathsf T},\\[-2pt]
V_t(\bm q)
 &= \bigl[\bm{p}_1(\bm q)^{\mathsf T}\ \cdots\
          \bm{p}_K(\bm q)^{\mathsf T}\ S^{\mathsf T}\bigr]^{\mathsf T}.
\end{aligned}
\end{equation}
A Delaunay tetrahedralization of $\widetilde V_t$ defines neighbor sets
$\mathcal N_{i,t}$ and the uniform Laplacian
$(L_tV)_i=\bm v_i-|\mathcal N_{i,t}|^{-1}
\sum_{j\in\mathcal N_{i,t}}\bm v_j$.
The interaction-mesh objective is
\vspace{-2pt}
\begin{equation}
 E_{\mathrm{mesh}}(\bm q)=
 \lVert L_tV_t(\bm q)-L_t\widetilde V_t\rVert_F^2.
 \label{eq:interaction_mesh}
\end{equation}

\textbf{TERRA residuals.}
TERRA augments the interaction-mesh objective with
segment- and terrain-interaction residual terms. The calibrated orientation residual
$\bm r_i^R=\Log(R_{i,t}^{\mathrm{src}}R_i(\bm q)^\top)$ matches source and
target-model rotations at the pelvis and bilateral hip and knee sites.
During stance, $\bm r_i^{xy}=\bm p_i^{xy}(\bm q)-\bm a_i$ holds each calcaneus at its touchdown
anchor $\bm a_i$. Ankle and toe-displacement terms penalize all
frame-to-frame horizontal motion and motion above $0.25$\,m/s.
One-sided residuals $[d_{\min}-d(\bm q)]_+$ and
$[h_t^\star-h_{\mathrm{sole}}(\bm q)]_+$ penalize proximity between left--right
leg collision geometries and insufficient swing-sole clearance, respectively.
Near terrain faces, vertical ankle and
toe targets add at most $0.06$\,m of terrain-clearing lift. Calibrated sole offsets align source landmarks with the robot's contact
surface. During detected seated rests, a signed-distance residual
brings the posterior pelvis collision geometries to the reconstructed seat. Detailed hyperparameter values are included in the Supp. Video.

\textbf{Sequential quadratic program.}
At iteration $n$ of frame $t$, we compute an update
$\Delta\bm q\in\mathbb R^{89}$ to the current configuration $\bm q_t^{(n)}$. The 89 generalized coordinates comprise three root translations, a unit
quaternion, and 82 joint coordinates. Let
$\mathcal B_t^{(n)}
=[\bm q^{\mathrm{lb}}-\bm q_t^{(n)},
  \bm q^{\mathrm{ub}}-\bm q_t^{(n)}]$:
\vspace{-2pt}
\begin{equation}
\label{eq:terra_qp}
\begin{aligned}
\Delta\bm q_t^{(n)}
=\argmin_{\Delta\bm q}\;&
 w_L\|\bm r_L+J_L\Delta\bm q\|_2^2
 +\|\Delta\bm q-\bm d_t\|_W^2\\[-2pt]
&+\|\bm q_t^{(n)}+\Delta\bm q\|_Q^2\\[-2pt]
&+\sum_{m\in\mathcal M_t}w_m
 \|J_m\Delta\bm q-\bm e_{m,t}\|_2^2\\[-2pt]
\mathrm{s.t.}\quad&
 \bar\phi_a+J_{\phi,a}\Delta\bm q
 \geq-\epsilon_{\mathrm{pen}},
 \quad a\in\mathcal C_t,\\[-2pt]
&\Delta\bm q\in
 \mathcal B_t^{(n)}\cap\eta\mathbb B_2 .
\end{aligned}
\end{equation}
Here
$\bm r_L=\operatorname{vec}\!\left(
L_tV_t(\bm q_t^{(n)})-L_t\widetilde V_t\right)$
is the current interaction-mesh residual and
$J_L=\partial\bm r_L/\partial\bm q$ is its Jacobian. Thus,
$\bm r_L+J_L\Delta\bm q$ approximates the residual after the update.
The vector
$\bm d_t=\bm q_{t-1}^{\star}-\bm q_t^{(n)}$ denotes the displacement from the current
iteration to the previous-frame solution. The matrix $W$ weights this temporal
smoothing residual, and $Q$ penalizes absolute trunk angles. The set
$\mathcal M_t$ constitutes the active TERRA residuals described above, and
$\mathcal C_t$ contains active
robot--environment geometry pairs. For
each pair $a$, $\phi_a$ is its signed separation distance, with $\phi_a<0$ denoting
penetration. We use the clipped distance
$\bar\phi_a=\max\{\phi_a,-(\epsilon_{\mathrm{pen}}+\rho)\}$ with
$\epsilon_{\mathrm{pen}}=0.9$\,mm and per-iteration recovery cap
$\rho=10$\,mm. Thus, an existing deep penetration requests at most 10\,mm of
correction in one SQP iteration rather than making the subproblem arbitrarily large. 
After each SQP solve, the root quaternion is radially projected onto the unit 3-sphere.
Native Clarabel is used, with the condensed CVXPY formulation as
a numerical fallback.

To warm-start the solution, we append 30 copies of the
first target frame and discard them afterward. The first solver frame uses trust
radius $1.0$ and at most 50 iterations; subsequent frames use radius $0.2$ and at
most 10 iterations. Finally, short collision outliers and tendon discontinuities are fixed via bounded interpolation after optimization.

\begin{figure}[!t]
  \centering
    \includegraphics[width=0.8\columnwidth]{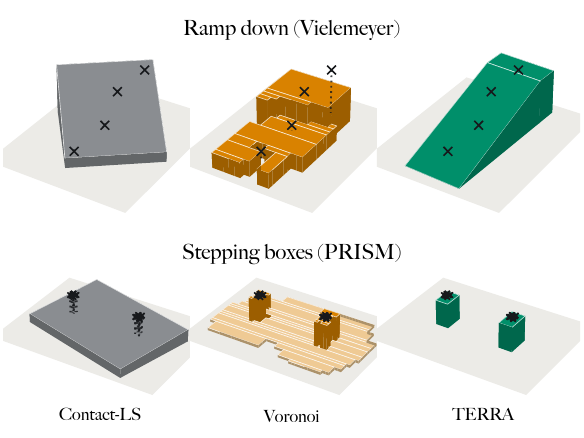}%
  \caption{\textbf{Terrain-reconstruction.} Method comparison on a Vielemeyer $10^\circ$
  descent ramp (top) and PRISM stepping boxes (bottom). Black crosses denote the
  same reference toe contacts in every panel, and dotted segments connect each contact
  to the reconstructed height at the same horizontal location. Vertical height is
  exaggerated for visibility.}
  \label{fig:terrain_reconstruction_ablations}
  \vspace{-15pt}
\end{figure}

\begin{table*}[!ht]
  \centering
  \scriptsize
  \setlength{\tabcolsep}{3.1pt}
  \caption{Terrain classification, reconstruction, and motion--terrain consistency
  across datasets. Entries are per-motion mean $\pm$ population standard deviation.}
  \label{tab:terrain_reconstruction}
  \vspace{-5pt}
  \begin{tabular}{lcccccc}
    \toprule
    \multicolumn{7}{c}{\textbf{(a) Known terrain}} \\
    & \shortstack{All known\\terrain} & \multicolumn{3}{c}{Gait120} & Darmstadt & Vielemeyer \\
    \cmidrule(lr){3-5}\cmidrule(lr){6-6}\cmidrule(lr){7-7}
    Method & \shortstack{Family acc. (\%) $\uparrow$\\($N=5{,}474$)}
      & \shortstack{Grade MAE ($^\circ$) $\downarrow$\\($N=1190$)}
      & \shortstack{Riser MAE (mm) $\downarrow$\\($N=1184$)}
      & \shortstack{Stool MAE (mm) $\downarrow$\\($N=1102$)}
      & \shortstack{Riser MAE (mm) $\downarrow$\\($N=1282$)}
      & \shortstack{Grade MAE ($^\circ$) $\downarrow$\\($N=568$)} \\
    \midrule
    Contact least squares  & N/A & \meanstd{1.683}{3.533} & \meanstd{23.42}{8.71}
      & \meanstd{490.00}{0.00} & \meanstd{150.49}{57.41} & \underline{\meanstd{1.199}{1.088}} \\
    Voronoi                & N/A & N/A & \meanstd{42.24}{39.54}
      & \underline{\meanstd{80.90}{121.19}} & \meanstd{24.81}{36.39} & N/A \\
    TERRA (without physical cues) & \underline{73.5} & \bestmeanstd{0.220}{0.235} & \underline{\meanstd{12.25}{5.35}}
      & \bestmeanstd{45.62}{13.57} & \underline{\meanstd{7.12}{6.10}}
      & \bestmeanstd{0.536}{0.333}\,(N=566) \\
    TERRA         & \textbf{99.9} & \underline{\meanstd{0.223}{0.244}} & \bestmeanstd{8.08}{6.99}
      & \bestmeanstd{45.62}{13.57} & \bestmeanstd{5.17}{4.36} & \bestmeanstd{0.536}{0.353} \\
    \bottomrule
  \end{tabular}
  \par\vspace{5pt}
  \setlength{\tabcolsep}{2.8pt}
  \begin{tabular}{lccccccc}
    \toprule
    \multicolumn{8}{c}{\textbf{(b) PRISM mesh geometry}} \\
    & & \multicolumn{3}{c}{Height MAE (mm) $\downarrow$}
      & \multicolumn{3}{c}{Footprint (\%) $\uparrow$} \\
    \cmidrule(lr){3-5}\cmidrule(lr){6-8}
    Method & Success & Foot ($N=18$) & Seat ($N=13$) & Observed ($N=31$) & Raised & Flat & IoU \\
    \midrule
    Contact least squares  & \textbf{31/31}
      & \meanstd{102.41}{34.09} & \meanstd{204.21}{19.14} & \meanstd{145.10}{57.89}
      & \bestmeanstd{99.1}{5.0} & \meanstd{37.1}{5.8} & \meanstd{3.5}{1.0} \\
    Voronoi                & \textbf{31/31}
      & \underline{\meanstd{34.13}{10.02}} & \underline{\meanstd{15.44}{8.38}} & \underline{\meanstd{26.29}{13.15}}
      & \underline{\meanstd{98.4}{4.6}} & \underline{\meanstd{40.3}{28.1}} & \underline{\meanstd{11.2}{17.9}} \\
    TERRA         & \textbf{31/31}
      & \bestmeanstd{12.95}{6.56} & \bestmeanstd{11.16}{5.93} & \bestmeanstd{12.20}{6.36}
      & \meanstd{90.8}{12.4} & \bestmeanstd{98.0}{0.6} & \bestmeanstd{47.3}{5.5} \\
    \bottomrule
  \end{tabular}
  \par\vspace{5pt}
  \setlength{\tabcolsep}{3.4pt}
  \begin{tabular}{lccccccc}
    \toprule
    \multicolumn{8}{c}{\textbf{(c) AMASS-terrain motion consistency}} \\
    Method & Success
      & \shortstack{Contact MAE\\(mm) $\downarrow$}
      & \shortstack{$|r_e|\leq50$ mm\\(\%) $\uparrow$}
      & \shortstack{Penetrating / floating\\(\%) $\downarrow$}
      & \shortstack{Raised P / R / F1\\(\%) $\uparrow$}
      & \shortstack{Toe--ankle pair MAE\\(mm) $\downarrow$}
      & \shortstack{Free-space penetration\\(\%) $\downarrow$} \\
    \midrule
    Contact least squares & \textbf{993/993} & \meanstd{47.46}{30.38}
      & 57.5 & 34.5 / 8.0 & 46.1 / \textbf{99.3} / 62.9 & \meanstd{24.73}{23.59} & \meanstd{0.225}{0.856} \\
    Voronoi & \underline{966/993} & \underline{\meanstd{25.57}{25.57}}
      & \underline{91.4} & \underline{2.0} / \underline{6.7} & \underline{80.0} / 82.2 / \underline{81.1} & \underline{\meanstd{19.04}{21.31}} & \bestmeanstd{0.000}{0.000} \\
    TERRA & \textbf{993/993} & \bestmeanstd{10.24}{9.21}
      & \textbf{97.1} & \textbf{0.7} / \textbf{2.2} & \textbf{98.6} / \underline{89.7} / \textbf{93.9}
      & \bestmeanstd{9.01}{7.46} & \underline{\meanstd{0.077}{0.444}} \\
    \bottomrule
  \end{tabular}
  \vspace{-5pt}
\end{table*}

\begin{table*}[!t]
  \centering
  \scriptsize
  \setlength{\tabcolsep}{2.0pt}
  \caption{Retargeting performance across datasets and held-out policy-tracking
  performance.}
  \vspace{-5pt}
  \label{tab:retargeting}
  \label{tab:retargeting_policy}
  \resizebox{\textwidth}{!}{%
  \begin{tabular}{lccccccccc}
    \toprule
    & \multicolumn{7}{c}{Retargeting} & \multicolumn{2}{c}{Policy tracking} \\
    \cmidrule(lr){2-8}\cmidrule(lr){9-10}
    Method & Success
      & $D_{\mathrm{ten}}$ (\%) $\downarrow$
      & $D_{\mathrm{col}}$ (\%) $\downarrow$
      & $D_{\mathrm{pen}}$ (\%) $\downarrow$
      & $D_{\mathrm{skat}}$ (\%) $\downarrow$
      & $D_{\mathrm{float}}$ (\%) $\downarrow$
      & \shortstack{Contact F1 (\%)\\$\uparrow$}
      & \shortstack{Success (\%)\\$\uparrow$}
      & \shortstack{MPJPE (mm)\\$\downarrow$} \\
    \midrule
    MM-MoCap-Body          & \underline{6478/6498} & \meanstd{2.208}{3.852} & \meanstd{12.09}{19.90} & \meanstd{54.56}{35.53} & \meanstd{4.86}{5.10} & \meanstd{22.80}{36.78} & \meanstd{92.86}{7.47} & \meanstd{47.34}{0.22} & \meanstd{90.26}{0.60} \\
    GMR              & \textbf{6498/6498} & \meanstd{0.556}{1.087} & \meanstd{21.82}{39.14} & \meanstd{19.98}{24.19} & \meanstd{9.25}{14.62} & \meanstd{11.82}{11.27} & \meanstd{74.19}{21.01} & \meanstd{56.43}{0.64} & \meanstd{91.97}{1.73} \\
    OmniRetarget     & 6392/6498 & \meanstd{0.466}{1.243} & \meanstd{4.98}{10.82} & \meanstd{31.97}{29.22} & \meanstd{12.70}{10.54} & \underline{\meanstd{3.61}{9.79}} & \meanstd{86.05}{11.79} & \meanstd{80.29}{1.39} & \meanstd{87.23}{2.92} \\
    TERRA (no residuals) & \textbf{6498/6498} & \meanstd{0.134}{0.407} & \meanstd{3.93}{8.45} & \meanstd{33.45}{33.63} & \meanstd{8.88}{7.03} & \meanstd{4.18}{11.00} & \meanstd{88.49}{9.51} & \underline{\meanstd{86.52}{0.25}} & \meanstd{80.58}{2.31} \\
    TERRA+Voronoi & 6470/6498 & \underline{\meanstd{0.081}{0.285}} & \underline{\meanstd{0.54}{1.35}} & \underline{\meanstd{2.99}{14.97}} & \underline{\meanstd{2.63}{2.82}} & \meanstd{9.97}{18.73} & \underline{\meanstd{92.96}{5.96}} & \meanstd{82.16}{0.41} & \bestmeanstd{71.65}{1.00} \\
    TERRA  & \textbf{6498/6498} & \bestmeanstd{0.026}{0.185} & \bestmeanstd{0.52}{1.34} & \bestmeanstd{0.16}{2.46} & \bestmeanstd{1.80}{1.87} & \bestmeanstd{1.02}{3.75} & \bestmeanstd{93.80}{4.06} & \bestmeanstd{87.59}{0.18} & \underline{\meanstd{74.47}{1.51}} \\
    \bottomrule
  \end{tabular}
  }
  \vspace{-15pt}
\end{table*}

\vspace{-3pt}
\subsection{Policy Training}
\label{sec:policy}

We trained multi-motion, terrain-aware, muscle-actuated control policies on terrain-paired motions via motion imitation. The tracking
policy observes joint positions and velocities, root height and projected gravity,
heading-frame root velocity, muscle commands and activation states, four foot and toe
touch sensors, and a heading-aligned $11\!\times\!11$ height map centered on the
pelvis at $0.1$\,m spacing. The one-step target encodes heading-frame root and
mimic-site errors in position, orientation, and velocity; site positions and
velocities are relative to the pelvis. Future targets at 0.2, 0.4, 0.6, and
0.8\,s encode root motion and pelvis-relative site positions without motion phase.

Training was performed with PPO on MJX--Warp~\cite{schulman2017proximal}.
The actor and critic are layer-normalized gated-residual MLPs. We used 8192 parallel
environments at a control frequency of 100\,Hz with five 2\,ms physics steps
per action.
We used adaptive motion sampling to increase the sampling likelihood of hard motions.

The PPO reward combines full-body pose and velocity tracking with a pelvis-relative
upper-body position term and a terminal quality bonus. Small
penalties discourage out-of-bounds and rapidly changing actions.
In addition, a muscle activity regularization term discourages excessive activation while preventing muscle-unit silencing.
Episodes terminate when
the global or core-upper-body mean site error exceeds $0.15$\,m or the root-orientation error exceeds $1$\,rad.
Each policy was trained for 4 billion steps. Hyperparameters are reported in Table~\ref{tab:implementation_specification}.

\begin{table}[!t]
  \centering
  \scriptsize
  \setlength{\tabcolsep}{3pt}
  \renewcommand{\arraystretch}{1.02}
  \caption{Key numerical parameters for reconstruction, retargeting, and policy training.}
  \label{tab:implementation_specification}
  \begin{tabular}{@{}p{0.27\columnwidth}p{0.67\columnwidth}@{}}
    \toprule
    Parameter & Value \\
    \midrule
    Stationary contacts
      & $v_s=0.30$\,m/s; $n_{\min}=5$; $W_e=\pm0.30$\,s;
      $\epsilon_{\rm loc}=0.05$\,m. \\
    Geometry margins
      & Contact expansion $0.10/0.05$\,m; maximum extension $0.00/0.15$\,m;
      free-space clearance $0.03/0.02$\,m (vertical/horizontal). \\
    TERRA weights
      & Orientation $0.5$; anchor $100$; velocity/displacement $150/200$; height $400$;
      separation $20{,}000$; clearance/route $1{,}500$; seat $2{,}000$. \\
    Constraint thresholds
      & $\epsilon_{\rm pen}=0.0009$\,m; recovery cap $\rho=0.010$\,m/iteration;
      leg separation $0.002$\,m; collision repair $0.005$\,m for $\leq3$ frames. \\
    $D_{\rm ten}$
      & Fraction with $\Delta l/l_0>\max(10\,\mathrm{EMA},0.001)$
      and $>0.05$. \\
    Actor/critic width
      & $2048,512,512,512,512,512,512$. \\
    PPO parameters
      & 8,192 envs; 20 steps; 1 epoch; 128 minibatches; $\gamma=0.99$;
      $\lambda=0.95$; clip 0.2; action SD 0.3; LR $8\!\times\!10^{-4}\rightarrow8\!\times\!10^{-5}$. \\
    Reward weights
      & Joint $q/v$: 0.5/0.25; root $p/R/v$: 1/0.5/0.5; rel. sites: 1/0.5/0.5;
      core 1; terminal 25; action rate $10^{-4}$; activation/floor 2/1. \\
    Training
      & 4B  steps; $\sim$20 hours on four A100 GPUs \\
    \bottomrule
  \end{tabular}
\end{table}

\section{Experimental Protocol}
\label{sec:experiments}

\textbf{Datasets.}
We sought to evaluate the full TERRA pipeline on a broad set of reference motions, combining diverse terrain interaction and biomechanical relevance. To do so, we applied TERRA on five distinct motion capture datasets. We extracted 993 non-flat motions from AMASS (AMASS-terrain)
for large-scale, diverse motion~\cite{mahmood2019amass}. We additionally considered 31 PRISM~\cite{hori2026grip} motions containing static terrain,
including ground-truth object meshes serving as
geometric references for terrain reconstruction evaluation.
Finally, we included 3,476 Gait120 motions ~\cite{boo2025gait120}, 1,282 Darmstadt stair
motions~\cite{grimmer2023darmstadt}, and 716 Vielemeyer ramp
motions~\cite{vielemeyer2026ramp} as openly available biomechanics datasets with known terrain geometry, enabling quantitative evaluation of terrain reconstruction accuracy and physiological comparison with human data. For policy experiments, we added 595 Gait120 level-walking clips and 975 AMASS flat motions inherited from Kinesis~\cite{simos2025kinesis}. Motion selection is described in detail in the Supplementary Video.

\begin{figure*}[!t]
  \centering
    \includegraphics[width=1\textwidth]{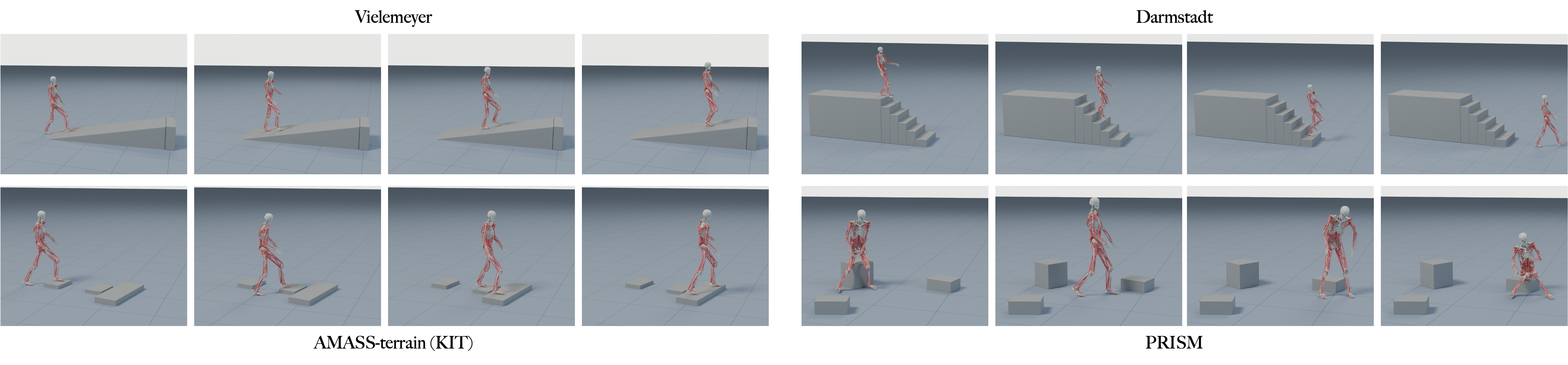}%
  \vspace{-10pt}
  \caption{A policy trained on TERRA motions can imitate held-out motion on various terrains, such as stairs, ramps, individual platforms, and chairs.}
  \label{fig:policy_examples}
  \vspace{-10pt}
\end{figure*}

\textbf{Reconstruction comparisons.}
To evaluate TERRA's terrain reconstruction performance, we compared it to two baselines: (i)~contact least squares, which fits one affine plane to the four-point kinematic contacts; and (ii)~Voronoi,
a height-field baseline adapted from TIP~\cite{jiang2022tip} and
SceneBot~\cite{chen2026scenebot}. We also ablated the effect of TERRA's swing-clearance cues to highlight their effect on terrain prior selection.
For a fair comparison on chair reconstruction, TERRA and Voronoi retained their own detection and
reconstruction but shared the posterior-mesh seat-height rule from
Sec.~\ref{sec:terrain}.

We evaluated all methods individually on every dataset, according to the available ground-truth information (Table~\ref{tab:terrain_reconstruction}). For Gait120, Darmstadt, and Vielemeyer, terrain labels and dimensions are known; we therefore report terrain-family accuracy and MAE in ramp grade, stair-riser height, and stool height. For PRISM, reference meshes additionally enable foot, seat, and combined-support height MAE, raised- and flat-region coverage, and raised-footprint intersection over union (IoU).
The family-accuracy denominator pools every labeled known-terrain clip:
$3{,}476$ Gait120 $+1{,}282$ Darmstadt $+716$ Vielemeyer $=5{,}474$;
the other columns are dataset-specific. Neither these labels nor apparatus dimensions
enter reconstruction. The numerical constants encode declared landmark, temporal, or
collision resolutions rather than nominal apparatus dimensions.

We report confidence intervals using subject-clustered bootstrap samples and break down reconstruction performance separately across terrain conditions. We also assessed robustness to noise (100-ms correlated isotropic Gaussian landmark noise at 2, 5, or 10 mm RMS), support interval removal (10\%, 20\%, and 30\%), and threshold perturbations (80\% and 120\% of the contact-speed, level-clustering, free-space, and minimum-raised-height).

Because AMASS lacks scene geometry, we evaluated motion--terrain consistency using
contact and non-collision criteria~\cite{hassan2021posa}. For event \(e\) and probe
\(j(e)\),
\begin{equation}
  r_e = h_m(x_e,y_e)-\left(z_e-\delta_{j(e)}\right),
  \label{eq:amass_consistency}
\end{equation}
where \(\delta_j\) is a source-only offset from the probe's lowest contact cluster.
We report contact-height MAE; consistency ($|r_e|\leq50$ mm), penetrating, and
floating event rates; raised-support precision/recall/F1 at 40 mm; toe--ankle
relative-support MAE; and non-support penetration below 30 mm. Continuous metrics
are unweighted per-motion mean $\pm$ population SD.

\begin{figure*}[!t]
  \centering
  \includegraphics[width=0.9\textwidth]{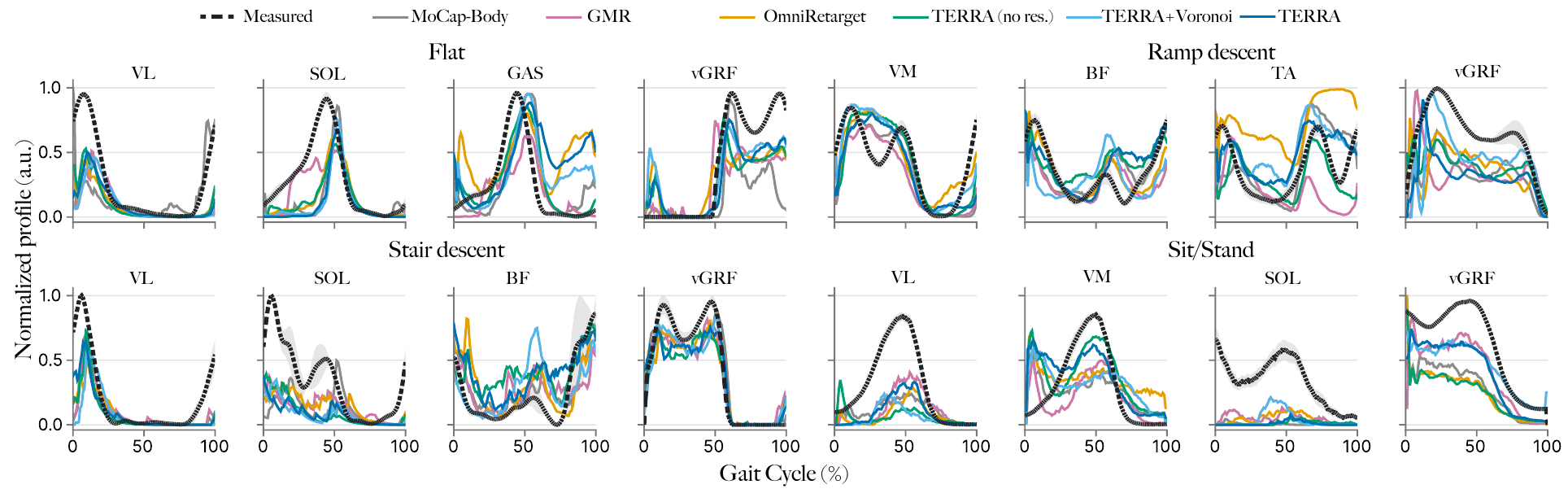}
  \vspace{-10pt}
  \caption{Qualitative comparison of population-averaged physiological profiles
  on held-out motions. Every human and policy trace was independently min--max normalized, yielding a waveform-shape
  comparison. Shading denotes between-subject SEM. VL: vastus lateralis; VM: vastus medialis;
  SOL: soleus; GAS: gastrocnemius; BF: biceps femoris; TA: tibialis anterior.}
  \label{fig:physiological-evaluation}
  \vspace{-20pt}
\end{figure*}

\textbf{Retargeting comparisons.}
We compared TERRA with OmniRetarget~\cite{yang2026omniretarget},
GMR~\cite{araujo2026gmr}, and MM-MoCap-Body~\cite{li2026musclemimic}.
Since these methods do not support terrain reconstruction, we used the same TERRA
scene across all baselines~(Table~\ref{tab:retargeting}). We used a 100\,Hz
retargeting framerate across all methods.
For OmniRetarget, we replaced the robot model with MyoFullBody,
rebuilt its bounds for all 82 scalar joint coordinates, and provided the fitted
SMPL-H shape, 17 landmark--body correspondences, and calibrated local offsets,
retaining the original interaction-mesh, object-nonpenetration, and foot-sticking
terms. For GMR, we followed the implementation from MuscleMimic~\cite{li2026musclemimic}, providing the actual terrain height for z-calibration. MM-MoCap-Body used MuscleMimic's native MyoFullBody fitter.
We furthermore ablated TERRA's residual terms to isolate their effect from the solver backend and other processing steps.
Finally, to highlight the effect of terrain reconstruction on downstream motion retargeting, we added a \emph{TERRA+Voronoi} baseline which combines Voronoi terrain reconstruction with TERRA's motion retargeting pipeline.

We define reference quality by preservation of task-relevant contacts together with satisfaction
of target-specific anatomical and environmental constraints. We evaluate source-contact
timing, tendon continuity, self-collision, penetration, skating, floating, and retargeting
coverage.
Coverage denotes converged retargeting runs over all motion inputs. We express each violation metric as a percentage of the relevant motion interval: the full motion interval for general constraints and source-derived contact or stance intervals for foot-specific constraints.
We report $D_{\rm ten}$ when an adaptive tendon-length event exceeds 0.05
relative rest length; skating when the corresponding robot ankle or toe moves horizontally
faster than $0.30$\,m/s, floating when a stance foot is more than $20$\,mm above its
assigned surface, inter-leg collision when penetration exceeds $1$\,mm, and
environment penetration when penetration exceeds $10$\,mm.
Contact preservation is the duration-weighted F1 score between source and retargeted
kinematic-contact labels over the left and right ankles and toes. We apply the same
contact detection criteria to both motions: probe speed below $0.30$\,m/s, height within
$50$\,mm of the local minimum in a $\pm0.30$\,s window, and a minimum contact duration
of $0.15$\,s. True positives, false positives, and false negatives are accumulated over
frame--probe duration, and $F_1=2\mathrm{TP}/(2\mathrm{TP}+\mathrm{FP}+\mathrm{FN})$.
Source labels are sampled at each method's declared output timestamps, with no temporal
shift or matching tolerance. Dataset results are pooled
by motion count and reported as the unweighted per-motion mean $\pm$ population
SD.

\textbf{Tracking policy comparisons.}
To assess downstream executability, we compared TERRA to its retargeting baselines by training a PPO policy from each method's retargeted references, using the common
setup (Sec.~\ref{sec:policy}). We partitioned the 8,068 source clips deterministically
by recorded identity (subject, or AMASS actor/session), targeting a 15\% test fraction
while balancing dataset and motion type.
For each method, we trained three independent policies with random seeds.
The comparison was performed on the held-out test motions successfully retargeted by all methods. We evaluated each motion clip with five stochastic runs.
Success denotes trajectory completion within a tracking termination bound of $0.25$m, and tracking error is average global mimic-site MPJPE over executed control steps.

\textbf{Physiological evaluation.}
We used held-out recordings from Gait120~\cite{boo2025gait120},
Darmstadt~\cite{grimmer2023darmstadt}, and Vielemeyer~\cite{vielemeyer2026ramp}
to visualize policy-generated EMG and vertical-GRF waveforms alongside human
data. Each case used ten stochastic
rollouts and a $0.25$-m termination threshold; every completed gait was retained,
including gaits from episodes that later terminated. Each completed policy gait
and each measured subject trace were independently min--max normalized to compare
waveform shape, focusing the analysis on temporal profiles. We averaged policy gaits within subject and then averaged policy
and measured traces separately across subjects in the same condition. We selected level
walking, ramp descent, stair descent, and sit/stand to span four task families with
available EMG and GRF profiles and, for each, show vertical GRF and
three representative muscles (Fig.~\ref{fig:physiological-evaluation}).

\section{Results}
\label{sec:results}

\subsection{Terrain Reconstruction}
Qualitatively, TERRA's priors enabled the method to flexibly infer slanted or piecewise-flat surfaces, whereas baselines methods were confined to more rigid representations (Fig.~\ref{fig:terrain_reconstruction_ablations}). We quantitatively evaluated the terrain reconstruction quality of TERRA by asking two key questions: how well does TERRA discriminate between different terrain types, and how closely does the inferred terrain geometry match the real geometry?
Overall, TERRA achieved 99.9\% terrain-family accuracy, successfully classifying ramps from stairs across datasets and conditions. Importantly, this classification accuracy was paired with strong reconstruction fidelity, namely sub-centimeter stair height accuracy, and sub-degree ramp accuracy, while outperforming baselines on all terrain categories
(Table~\ref{tab:terrain_reconstruction}). 
Removing the foot-orientation and swing-clearance cues reduced family accuracy to 73.5\%, while having little effect on other metrics (performance is identical to TERRA for Tables ~\ref{tab:terrain_reconstruction}b,c and thus not shown).

The subject-clustered 95\% CI for pooled family accuracy was 99.853--99.982\%
(144 dataset--subject clusters). Across independent captures, grade MAE was
0.544$^\circ$/0.529$^\circ$ at 7.5$^\circ$/10$^\circ$, and riser MAE was
7.36/4.34/3.61 mm at 100/170/240 mm.
TERRA remained robust to threshold
perturbations and landmark noise, while substantial support removal caused a gradual,
terrain-dependent degradation, most notably for stools (Table~\ref{tab:reconstruction_robustness}).

\begin{table}[!t]
  \centering
  \caption{Terrain reconstruction accuracy (\%) across perturbations.}
  \vspace{-10pt}
  \label{tab:reconstruction_robustness}
  \scriptsize
  \setlength{\tabcolsep}{2.5pt}
  \begin{tabular}{@{}lrrrrr@{}}
    \toprule
    & Level & Ramp & Stairs & Stool & All \\
    & (148) & (1,758) & (2,466) & (1,102) & (5,474) \\
    \midrule
    Clean & 100.00 & 100.00 & 99.84 & 100.00 & 99.93 \\
    Noise 2 mm & 100.00 & 99.98 & 99.81 & 100.00 & 99.91 \\
    Noise 5 mm & 100.00 & 99.75 & 99.10 & 100.00 & 99.51 \\
    Noise 10 mm & 99.59 & 95.78 & 96.52 & 99.96 & 97.06 \\
    Support $-10\%$ & 100.00 & 98.76 & 99.61 & 89.46 & 97.30 \\
    Support $-20\%$ & 100.00 & 96.64 & 98.96 & 79.84 & 94.40 \\
    Support $-30\%$ & 100.00 & 91.18 & 97.42 & 70.36 & 90.04 \\
    Thresholds $80\%$ & 100.00 & 99.89 & 99.96 & 100.00 & 99.95 \\
    Thresholds $120\%$ & 100.00 & 100.00 & 99.80 & 100.00 & 99.91 \\
    \bottomrule
  \end{tabular}
  \vspace{-13pt}
\end{table}

\subsection{Motion Retargeting}
We tested TERRA and its baselines on source-contact preservation as well as anatomical
and terrain-interaction validity. TERRA achieved the best results on all five physical
violation metrics and contact-timing F1 (Table~\ref{tab:retargeting}). TERRA's residual optimization
terms raised contact-timing F1 from 88.49\% to 93.80\% while lowering all
five violation metrics---especially terrain penetration, which dropped from 33.45\% to
0.16\%. Compared with TERRA+Voronoi, the reconstructed TERRA terrain further reduced
penetration, skating, and floating while increasing Contact F1.

\vspace{-5pt}
\subsection{Tracking Policy Performance}
\label{results:policies}
Qualitatively the policy performed well across different affordances (Fig.~\ref{fig:policy_examples}). We next asked whether improvements in reference validity were accompanied by improved downstream imitation performance. In the held-out evaluation setting, TERRA-retargeted reference motions produced the highest
completion rate compared to baselines
(Table~\ref{tab:retargeting_policy}); TERRA+Voronoi achieved better MPJPE but with a lower success rate. Broken down by terrain family,
TERRA achieved the highest success rate on flat ground, ramps, boxes, and seats,
and the lowest MPJPE on flat ground, ramps, and boxes
(Table~\ref{tab:policy_tracking_terrain}), while TERRA+Voronoi performed best on
stairs and attained the lowest seat MPJPE. We encourage readers to view the supplementary video for in-depth qualitative results.

\begin{table}[!t]
  \centering
  \scriptsize
  \setlength{\tabcolsep}{3.2pt}
  \caption{Held-out policy tracking by terrain family.}
  \vspace{-10pt}
  \label{tab:policy_tracking_terrain}
  \begin{tabular}{@{}lccc@{}}
    \toprule
    Terrain ($N$) & OmniRetarget & TERRA+Voronoi & TERRA \\
    \midrule
    \multicolumn{4}{@{}l}{\textit{Success rate (\%)}} \\
    Flat (262) & \meanstd{88.07}{0.62} & \meanstd{91.09}{0.75} & \bestmeanstd{91.70}{0.46} \\
    Ramp (286) & \meanstd{98.09}{0.08} & \meanstd{89.18}{1.06} & \bestmeanstd{99.49}{0.18} \\
    Stairs (422) & \meanstd{74.14}{3.49} & \bestmeanstd{85.01}{0.24} & \meanstd{81.90}{1.73} \\
    Boxes (61) & \meanstd{38.91}{4.64} & \meanstd{39.45}{0.19} & \bestmeanstd{46.45}{4.45} \\
    Seats (246) & \meanstd{72.11}{0.61} & \meanstd{70.19}{0.38} & \bestmeanstd{89.32}{0.82} \\
    \midrule
    \multicolumn{4}{@{}l}{\textit{MPJPE (mm)}} \\
    Flat (262) & \meanstd{82.42}{3.13} & \bestmeanstd{69.97}{1.57} & \meanstd{70.01}{2.55} \\
    Ramp (286) & \meanstd{75.82}{3.36} & \meanstd{67.83}{1.02} & \bestmeanstd{66.74}{2.92} \\
    Stairs (422) & \meanstd{85.18}{3.06} & \bestmeanstd{66.65}{1.12} & \meanstd{69.50}{2.57} \\
    Boxes (61) & \meanstd{94.02}{2.26} & \meanstd{85.25}{0.55} & \bestmeanstd{82.50}{0.77} \\
    Seats (246) & \meanstd{107.45}{2.48} & \bestmeanstd{83.09}{1.13} & \meanstd{94.75}{2.50} \\
    \bottomrule
  \end{tabular}
  \vspace{-10pt}
\end{table}

\subsection{Physiological Evaluation}
Terrain-paired biomechanics datasets let us compare signals produced by learned
policies directly with measured human profiles for the same locomotion condition.
Across the selected held-out flat, ramp, stair, and sit/stand examples, the policy and
human profiles show both similarities and visible mismatches in EMG timing and shape
(Fig.~\ref{fig:physiological-evaluation}).
The vertical-GRF panels provide a visual comparison of stance timing and normalized
waveform shape. Taken together, these examples suggest that TERRA can serve as a tool to study physiological processes in the future. 

\section{Conclusion}
\label{sec:conclusion}

TERRA demonstrates that motion kinematics can provide enough environmental
evidence to recover task-relevant collision geometry for muscle-actuated control.

Across stairs, ramps, isolated supports, and chairs, TERRA reduced anatomical and interaction violations
and yielded the highest held-out completion rate.
By incorporating diverse terrain navigation into the behavioral repertoire of musculoskeletal control policies, TERRA may open a novel route to studying affordance representation in embodied agents.

For instance, classical findings that stair climbing and sitting are perceived in body-scaled terms~\cite{warren1984perceiving,mark1987eyeheight} could be revisited by relating learned terrain representations to the musculoskeletal capabilities of the body, complementing neural accounts of how the brain specifies competing action opportunities~\cite{cisek2010neural,angelaki2026simons}.
Dynamic primitive
geometry, online retargeting, and moving surfaces remain as open challenges for future work.




\end{document}